\documentclass{article}

\usepackage[accepted]{icml2025}

\usepackage{microtype}
\usepackage{graphicx}
\usepackage{subfigure}
\usepackage{booktabs} %
\usepackage{comment}
\usepackage{hyperref}
\usepackage{xcolor}
\usepackage{subcaption}

\usepackage{amsmath}
\usepackage{amssymb}
\usepackage{mathtools}
\usepackage{amsthm}

\usepackage[capitalize,noabbrev]{cleveref}

\theoremstyle{plain}

\theoremstyle{definition}

\theoremstyle{remark}

\usepackage[disable,textsize=tiny]{todonotes}

\usepackage{tikz}
\usetikzlibrary{shapes.geometric, arrows}

\tikzstyle{startstop} = [rectangle, rounded corners, minimum width=3cm, minimum height=1cm,text centered, draw=black, fill=red!30]
\tikzstyle{process} = [rectangle, minimum width=3cm, minimum height=1cm, text centered, draw=black, fill=blue!30]
\tikzstyle{arrow} = [thick,->,>=stealth]
\tikzstyle{data} = [ellipse, minimum width=3cm, minimum height=1cm, text centered, draw=black, fill=green!30]
\tikzstyle{decision} = [diamond, minimum width=3cm, minimum height=1cm, text centered, draw=black, fill=yellow!30]

\usepackage[english]{babel}

\newcommand{\KLOM}{KLoM}
\newcommand{\ULIRA}{U-LiRA}

\newcommand{\GUS}{GUS}
\newcommand{\TOFU}{TOFU}
\newcommand{\WMDP}{WMDP}

\begin{document}
\twocolumn[
\icmltitle{Easy \textbf{D}ata \textbf{U}nlearning \textbf{B}ench}
\icmlsetsymbol{equal}{*}
\begin{icmlauthorlist}
\icmlauthor{Roy Rinberg\textsuperscript{*} \textsuperscript{§}}{harvard}
\icmlauthor{Pol Puigdemont\textsuperscript{*}}{epfl}
\icmlauthor{Martin Pawelczyk}{harvard}
\icmlauthor{Volkan Cevher}{epfl}
\end{icmlauthorlist}

\icmlaffiliation{epfl}{LIONS, EPFL, Lausanne, Switzerland}
\icmlaffiliation{harvard}{Harvard University}

\icmlcorrespondingauthor{Roy Rinberg}{royrinberg@g.harvard.edu}
\vskip 0.15in
\begin{center}
\href{https://github.com/easydub/EasyDUB-code}{\raisebox{-0.3\height}{\includegraphics[height=1.2em]{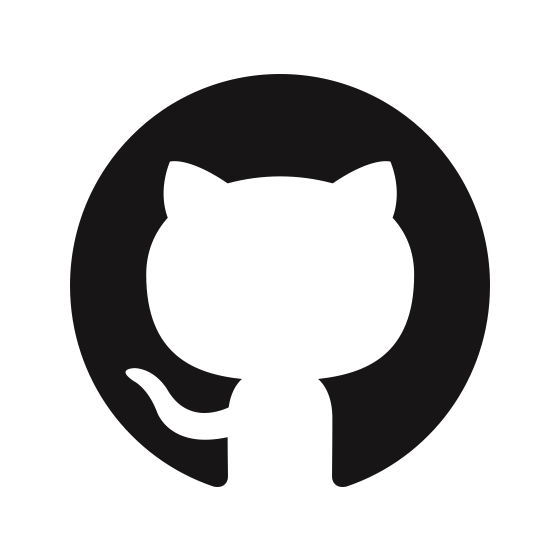}}\hspace{0.3em}Code}
\hspace{1.5em}
\href{https://huggingface.co/datasets/easydub/EasyDUB-dataset}{\raisebox{-0.3\height}{\includegraphics[height=1.2em]{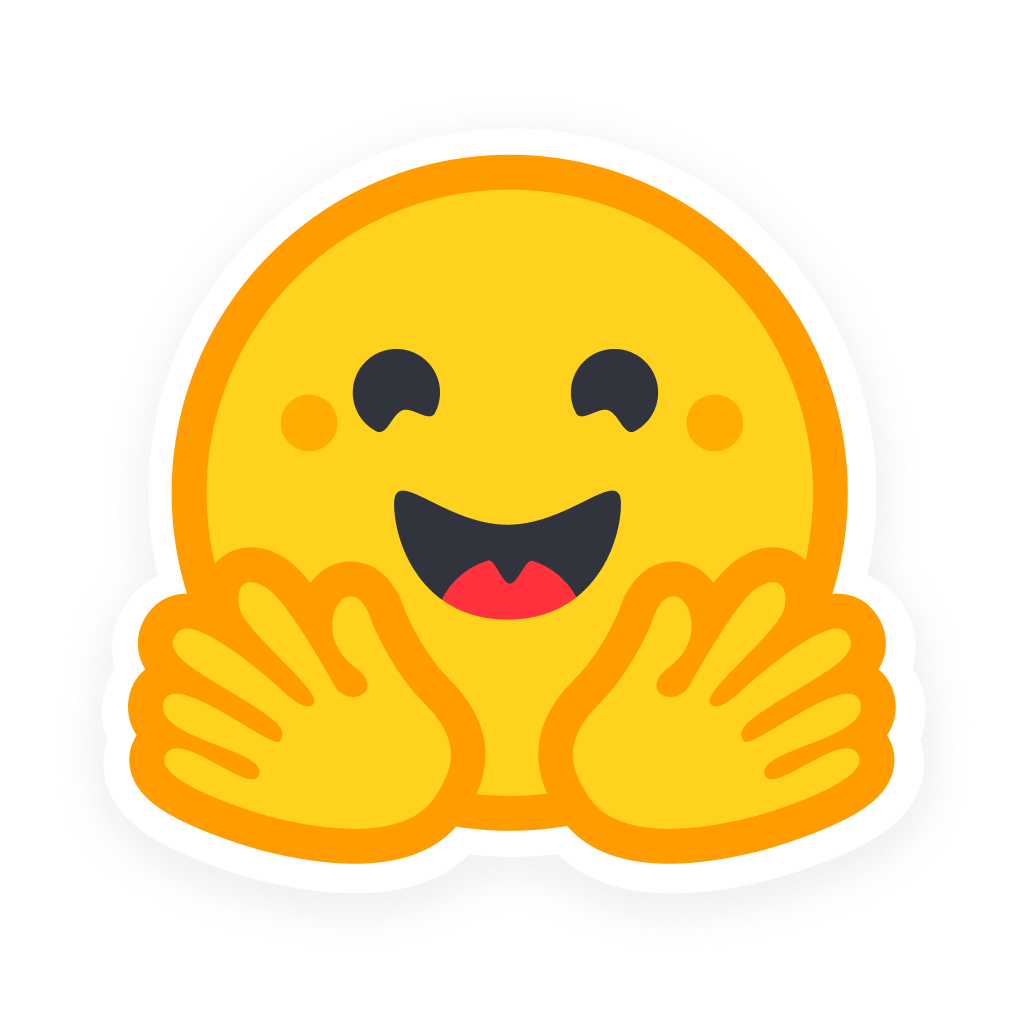}}\hspace{0.3em}Dataset}
\end{center}
\vskip 0.3in
]

\printAffiliationsAndNotice{\textsuperscript{*} Equal contribution. \textsuperscript{§} RR was funded in part by NSF grant BCS-2218803 (HNDS-I) and the Digital Data Design Institute at Harvard.}

\begin{abstract}
Evaluating machine unlearning methods remains technically challenging, with recent benchmarks requiring complex setups and significant engineering overhead. We introduce a unified and extensible benchmarking suite that simplifies the evaluation of unlearning algorithms using the \KLOM{} (KL divergence of Margins) metric \cite{georgiev2024attributetodeletemachineunlearningdatamodel}. Our framework provides precomputed model ensembles, oracle outputs, and streamlined infrastructure for running evaluations out of the box. By standardizing setup and metrics, it enables reproducible, scalable, and fair comparison across unlearning methods. We aim for this benchmark to serve as a practical foundation for accelerating research and promoting best practices in machine unlearning. Our \href{https://github.com/easydub/EasyDUB-code}{code} and \href{https://huggingface.co/datasets/easydub/EasyDUB-dataset}{data} are publicly available.

\end{abstract}

\section{Introduction}

The growing reliance on machine learning in sensitive and regulated domains, such as healthcare and finance, has raised critical concerns regarding the ability of models to selectively forget training data upon request. This process is known as \textit{machine unlearning}. Effective machine unlearning ensures that an unlearned model trained with certain data behaves as though specific data (the \textit{forget set}) was never included in the training set. Although retraining from scratch without the unwanted data offers a theoretically perfect solution, it remains computationally impractical, particularly for large-scale deep learning models.

Recent literature has introduced various heuristic methods aimed at approximate unlearning, yet rigorously evaluating these methods remains challenging. A significant barrier is the substantial computational cost involved in hyperparameter searches and repeated retrainings necessary to obtain reliable evaluation results. Current benchmarks are often complex, opaque, or costly, leading researchers to either spend extensive resources on evaluations or rely on oversimplified heuristics. This complexity can inadvertently promote evaluation methods that are susceptible to "gaming," where improved scores do not necessarily reflect genuine unlearning efficacy.

Motivated by these challenges, we propose a unified benchmarking framework that simplifies the evaluation of data unlearning methods. Our framework provides standardized infrastructure and readily available resources, including precomputed model ensembles, oracle outputs, and established evaluation protocols. Central to our evaluation is the \KLOM{} (KL divergence of Margins) metric, which quantifies the similarity of an unlearned model's predictive margins to those of an oracle model retrained without the target data \citep{georgiev2024attributetodeletemachineunlearningdatamodel}. By offering these resources publicly, our goal is to significantly lower the computational overhead of evaluations and to encourage the development of efficient heuristic approximations for unlearning metrics.

Beyond immediate computational efficiency, our benchmark also facilitates deeper investigation into fundamental aspects of machine unlearning, such as scaling laws related to model sizes and the transferability of unlearning across different data subsets. Furthermore, we outline future extensions, such as incorporating complementary evaluation metrics like the Gaussian Unlearning Score (GUS) \citep{pawelczyk2024machine} and addressing limitations observed in existing synthetic forgetting tasks such as TOFU \citep{maini2024tofutaskfictitiousunlearning}.

Through standardized, reproducible, and efficient evaluation of unlearning methods, we hope our benchmark accelerates progress towards practical, reliable, and computationally efficient unlearning methods, supporting more robust and responsible machine learning.

\vspace{-10pt}
\paragraph{Scope.}
While our benchmark focuses on classification models evaluated via predictive margins under the logistic loss, its insights extend to generative models. These models, including large language models (LLMs), are trained with cross-entropy loss, which decomposes into conditional classification tasks. The evaluation methods we propose are conceptually aligned with the unlearning challenges in generative modeling. Our work provides tools that can support future efforts to evaluate unlearning in generative AI.

\paragraph{Acknowledgements.}
This benchmark builds on tools, methodology, and insights first developed in the \emph{Attribute to Delete} study \citep{georgiev2024attributetodeletemachineunlearningdatamodel}, which laid the groundwork for using margin-based divergences as a principled evaluation framework for machine unlearning. We thank the authors for their foundational contributions.

\section{Metrics}

\subsection{Problems with Existing Metrics: \ULIRA{}}

We recall that the goal of data machine unlearning is to produce models whose behavior closely matches that of models retrained without the forget set. Achieving this requires statistical closeness between the distributions of the unlearned model and an oracle model, defined as the model retrained from scratch without the forget set. However, directly evaluating this objective has proven challenging, prompting the development of empirical proxies like \ULIRA{} \citep{hayes2024inexact}, which assess unlearning quality via adversarial distinguishability.

\ULIRA{} leverages membership inference attacks \citep{membership2022carlini} to measure if an adversary can discern whether a data point originated from the forget set or a held-out validation set based solely on the model’s output. Ideally, responses from the unlearned model to both sets would be indistinguishable, implying that an adversary cannot perform better than random guessing.

Despite its intuitive appeal, \ULIRA{} has several limitations. First, it is susceptible to manipulation. A trivial strategy to achieve a perfect \ULIRA{} score involves outputting constant margins across all inputs, rendering outputs indistinguishable. While this guarantees perfect indistinguishability, it completely eliminates model utility. This highlights a structural vulnerability in \ULIRA{}: it prioritizes indistinguishability without enough enforcing of utility constraints. Consequently, algorithms optimized for \ULIRA{} might inadvertently collapse model functionality, misleadingly inflating performance metrics.

Second, \ULIRA{} does not align fully with the formal definition of unlearning. The $(\varepsilon, \delta)$-unlearning criterion demands that the full distribution of model outputs, not merely distinguishability at specific points, must closely match the oracle distribution.

Another issue with \ULIRA{} is treating the forget set itself. The metric evaluates performance by randomly generating multiple forget sets and comparing models that unlearned specific points against models that unlearned other points. While this approach captures some variations, it overlooks potential "compound effects" arising from the particular composition of a given forget set. Specifically, the impact of removing one point may depend critically on the removal of other points simultaneously. Such nuanced interactions are obscured when forget sets vary randomly across evaluations. In contrast, methods like \KLOM{} address this limitation by fixing the forget set, thus providing a clearer understanding of compound interactions among points.

In summary, although \ULIRA{} provides a useful heuristic for evaluating unlearning, its susceptibility to manipulations and its scope limitations relative to the formal definition of unlearning indicate that caution is necessary when interpreting results. Metrics that directly estimate distributional divergence from oracle models inherently consider utility and are better suited for rigorous evaluation of unlearning quality.

\subsection{\KLOM{}}
\label{sec:klom}
The KL divergence of margins \citep{georgiev2024attributetodeletemachineunlearningdatamodel} (\KLOM{}) is a metric for empirically assessing machine unlearning. It measures how similar the output distributions of unlearned models are to those of oracle models, which are retrained from scratch without the forget set. \KLOM{} implements a relaxed version of the standard $(\varepsilon, \delta)$ unlearning definition by substituting KL divergence for approximate max divergence and by comparing model outputs instead of parameters.

To compute \KLOM{}, we first define the margin. For an input $x$ with true label $y_x$, let $f(x;\theta) \in \mathbb{R}^K$ be the output logits produced by a model $\theta$. The margin of model $\theta$ on input $x$ is defined as:
\[ \varphi(x; \theta) = (f(x;\theta))_{y_x} - \log \sum_{k \ne y_x} \exp((f(x;\theta))_k). \]
This logit-gap formulation captures how separable the correct prediction is from alternatives and is both numerically stable and aligned with our implementation.

\begin{figure}[ht]
  \centering
  \includegraphics[width=\linewidth]{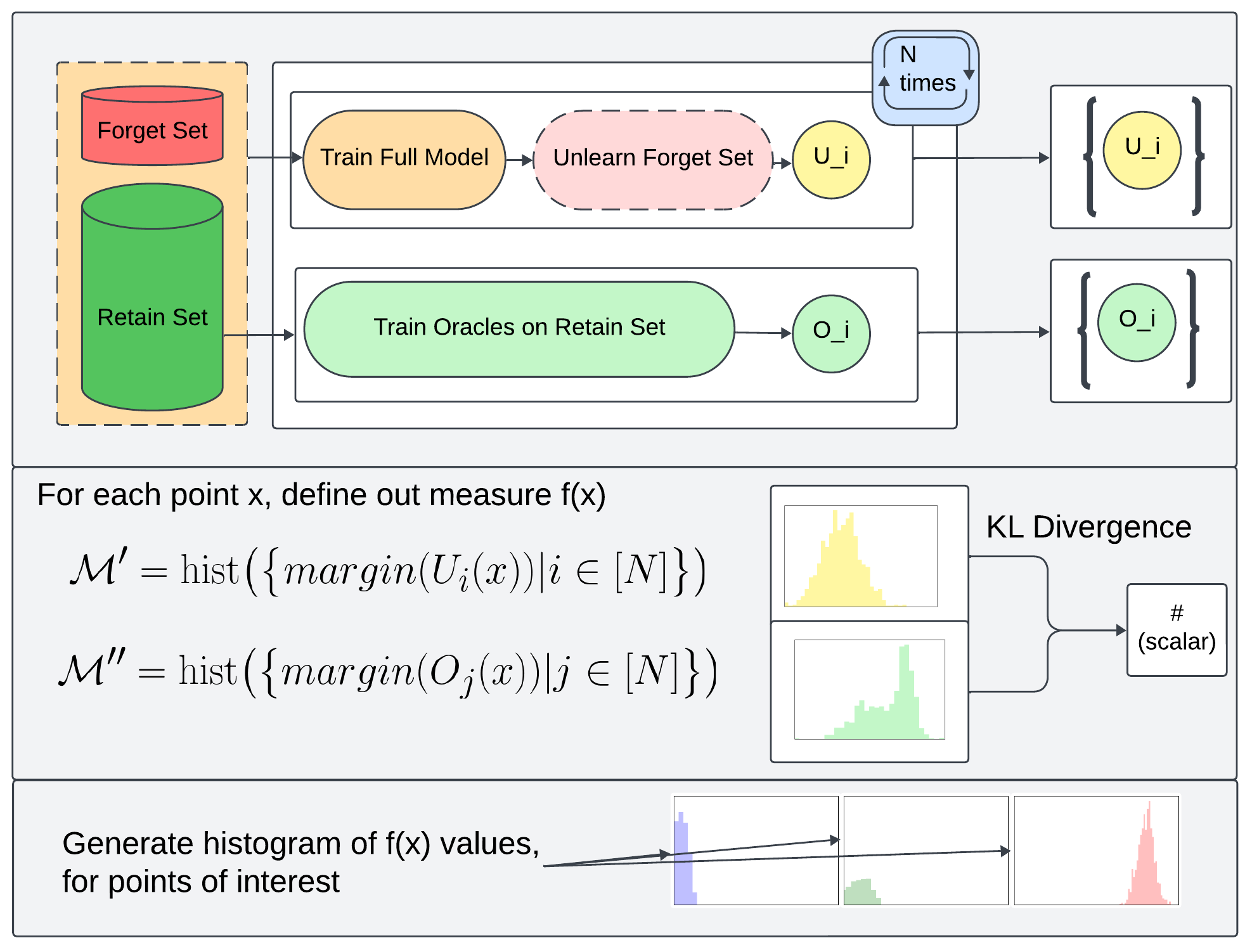}
  \caption{Overview of the \KLOM{} methodology \citep{georgiev2024attributetodeletemachineunlearningdatamodel}.}
  \label{fig:klom_methodology}
\end{figure}

We then compare margin distributions from (i) oracle models $\{\theta^o_i\}_{i=1}^N$ (trained on the retain set $S \setminus F$) and (ii) unlearned models $\{\theta^f_i\}_{i=1}^N$ (obtained by applying unlearning algorithm $U$ to models originally trained on the full dataset $S$). For each data point $x$, we compute its margin under each oracle and unlearned model, yielding two empirical margin distributions: $\{\varphi(x; \theta^o_i)\}_{i=1}^N$ and $\{\varphi(x; \theta^f_i)\}_{i=1}^N$. These are binned into histograms, and their KL divergence yields the pointwise metric:
{\small %
\[
\text{\KLOM{}}(x) = D_{\text{KL}}(\text{Hist}(\{\varphi(x; \theta^o_i)\}_{i=1}^N) \, \| \, \text{Hist}(\{\varphi(x; \theta^f_i)\}_{i=1}^N))
\]
} %

which can be averaged or aggregated across points in the forget, retain, or validation sets. Figure~\ref{fig:klom_methodology} provides a visual overview of this process.

\vspace{-0.5em}
\paragraph{Assumptions and Robustness}

\KLOM{} assumes only that we can draw repeated samples from the oracle and unlearned model distributions and observe their outputs. Unlike other metrics such as U-LiRA, it does not rely on Gaussian approximations or adversarial distinguishers. It is robust to “gaming” by degenerate unlearning strategies (e.g., returning a random model), which are penalized by KL divergence against the oracle distribution. This makes \KLOM{} harder to fool than membership-inference-based metrics.

\vspace{-10pt}
\paragraph{Hyperparameters and Practical Estimation}

\KLOM{} depends on several hyperparameters: (i) the number of models $N$, set to $100$ oracle and $100$ unlearned models for stable estimation (our dataset provides up to 200 of each, though $N{=}100$ suffices; see Appendix~\ref{app:klom_sensitivity}); (ii) a clipping range of $[-100, 100]$ to suppress extreme margin values from unstable models; (iii) the number of histogram bins, fixed at $20$ to balance resolution and variance; and (iv) a smoothing constant $\varepsilon = 10^{-5}$ to prevent empty histogram bins, which limits \KLOM{} values to a maximum of approximately $12$. We find the default parameter selection to be suitable for a fair evaluation across methods. In Appendix \ref{app:klom_sensitivity} we show \KLOM{} scores versus the number of compared models $N$ and conclude that $100$ models is enough for a reliable evaluation.

\subsection{Future Directions: Extending to LLMs}
\label{sec:extending-to-llms}
Some current LLM unlearning evaluations for unlearning, such as \TOFU{} \citep{maini2024tofutaskfictitiousunlearning} and Weapons of Mass Destruction Proxy (\WMDP{}) benchmark \citep{li2024wmdp}, rely on multiple-choice question (MCQ) answering. While very convenient, this format fails to test whether the model has truly forgotten information. For instance, a model can learn to detect sensitive topics and produce generic or misleading outputs without necessarily removing the underlying knowledge \citep{qi2024safetyalignmentjusttokens}. This evades detection while preserving the data internally. Robust evaluation should go beyond surface accuracy and measure whether the model’s output distribution matches that of a retrained oracle. Without this, current methods risk overstating unlearning success.

\paragraph{Teacher-forcing \KLOM{}}

We propose to extend \KLOM{} to language models by evaluating margin divergence under teacher forcing. For a token sequence $x = (w_1, \dots, w_T)$ and a model $\theta$, we compute the margin $\varphi_t(x; \theta)$ at each prediction step $t$. This margin represents the model's confidence in the true next token $w_{t+1}$ (from sequence $x$) relative to alternatives, given the prefix $x_{<t+1}=(w_1, \dots, w_t)$. It is computed using the logit-gap definition analogous to $\varphi(x'; \theta)$ in Section~\ref{sec:klom}, where $x'$ corresponds to the input context (prefix $x_{<t+1}$) and the 'true label' is $w_{t+1}$.

At each prediction step $t$, we compare margin histograms from oracle $\{\theta^o_i\}_{i=1}^N$ and unlearned $\{\theta^f_i\}_{i=1}^N$ model ensembles. For a sequence $x$, we generate margin sets $\{\varphi_t(x; \theta^o_i)\}_{i=1}^N$ and $\{\varphi_t(x; \theta^f_i)\}_{i=1}^N$. Let $\mathrm{Hist}^o_t(x)$ and $\mathrm{Hist}^f_t(x)$ be histograms from these respective sets. Then $\KLOM_t(x) = D_{\mathrm{KL}}(\mathrm{Hist}^o_t(x) \,\|\, \mathrm{Hist}^f_t(x))$. The teacher-forcing \KLOM{} is the average across token positions, $\overline{\KLOM}(x) = \frac{1}{T} \sum_{t=1}^T \KLOM_t(x)$, aggregated over a dataset $\mathcal{D}$ as $\overline{\KLOM}(\mathcal{D}) = \frac{1}{|\mathcal{D}|} \sum_{x \in \mathcal{D}} \overline{\KLOM}(x)$.

This formulation preserves the original metric's robustness while enabling distributional comparison for next-token predictions in autoregressive models. It requires no changes to \KLOM{} hyperparameters and is fully compatible with standard teacher-forcing evaluation.

Crucially, teacher-forcing \KLOM{} is significantly harder to game than multiple-choice formats. Rather than checking if a model avoids specific outputs, it compares the full predictive distribution against that of a ground truth model, capturing subtler forms of retained knowledge.

\section{Benchmark}
\label{sec:benchmark}
Evaluating machine unlearning requires comparing the predictions of unlearned models to those of oracle models retrained without the forget set. Doing so reliably demands generating ensembles of pre-trained and oracle models, computing classification margins for each, and then estimating divergence metrics such as \KLOM{}. Establishing such a reliable evaluation from scratch induces substantial overhead: for each forget set, one would typically need to train $N$ full-data models and $N$ oracles on the retain set, and extract per-example margin distributions from both. The cost of this setup is significant, both computationally and in engineering effort, presenting a bottleneck for rapid development and fair comparison of new methods.

To address this, our benchmark is designed with three key principles: (i) \textit{Reusable infrastructure:} Pre-trained and oracle model ensembles are agnostic to the unlearning method under test and are expensive to compute. We precompute and distribute these components, allowing users to focus solely on the unlearning algorithm. (ii) \textit{Standardized evaluation:} The benchmark provides tested implementations of core evaluation routines, reducing the risk of methodological errors and improving reproducibility. Users can trust that results are measured under consistent conditions. (iii) \textit{Turnkey experimentation:} A complete experimental pipeline supports automatic dataset download from HuggingFace, model loading, checkpointing, and end-to-end evaluation.

\paragraph{Included Datasets and Resources}

We provide full experimental support for the CIFAR-10 \citep{krizhevsky2009learning} benchmark from \citet{georgiev2024attributetodeletemachineunlearningdatamodel} using ResNet-9 models. The dataset ships with 200 pre-trained models trained on the full dataset, 10 distinct forget sets (ranging from 10 to 1{,}000 samples, including random, PCA-based, and CLIP-based selections), each with 200 corresponding oracle models retrained on the retain set, and precomputed classification margins for all models across validation, forget, and retain sets. In total the benchmark includes 2{,}200 model checkpoints and their precomputed margins, hosted on HuggingFace and automatically downloaded on first use.

\paragraph{Quick Start}
The dataset is hosted on HuggingFace and downloads automatically on first use. A complete experiment, from unlearning to \KLOM{} evaluation, runs with a single command:

\begin{verbatim}
uv run python demo.py \
  --method noisy_descent \
  --forget-set 1 --n-models 100
\end{verbatim}

This runs noisy-SGD unlearning on 100 pretrain models for forget set~1 and prints \KLOM{} scores against the oracle ensemble. No separate install or dataset download step is needed. Users can also compute \KLOM{} directly from precomputed margins without running any unlearning or requiring a GPU (see Appendix~\ref{app:python_api}).

The benchmark is designed to be easily extensible. To add a new unlearning method, users define a function in \texttt{unlearning.py} following the standard interface and register it in the \texttt{UNLEARNING\_METHODS} dictionary. The framework handles model loading, margin computation, and \KLOM{} evaluation automatically. A typical implementation requires fewer than 50 lines of code.

\section{Conclusions and Future Work}
\label{sec:conclusions}
We collected a benchmark based on \citep{georgiev2024attributetodeletemachineunlearningdatamodel} that makes evaluating machine unlearning both practical and rigorous. By providing precomputed model ensembles, oracle outputs, and tested evaluation tools, our framework allows researchers to compare unlearning methods under standardized and reproducible conditions. The infrastructure is designed to be easy to use and supports reliable evaluation with strong baselines, such as retraining without the forget set. Our main evaluation metric, \KLOM{}, offers a clear and tractable way to measure how close unlearned model predictions are to those of oracle models. We believe this benchmark lowers the barrier to entry and will help accelerate progress in developing more effective and efficient data unlearning algorithms along with better comparing of more efficient heuristic evals.

This benchmark can be extended in several promising directions. First, we are working on incorporating the Gaussian Unlearning Score (\GUS{}) \citep{pawelczyk2024machine} unlearning evaluation, which introduces a complementary perspective by quantifying unlearning efficacy through data poisoning reversibility. In a similar line, we are interested in extending our evaluations to support LLMs more broadly. Improving upon other current methods evaluated on synthetic forget tasks, such as TOFU \citep{maini2024tofutaskfictitiousunlearning}, and in understanding how unlearning techniques can scale to generative settings. We provided the initial discussion in Section~\ref{sec:extending-to-llms}. We also consider the introduction of a public leaderboard, similar in spirit to \textsc{RobustBench} \citep{croce2020robustbench}, to facilitate community engagement and transparent reporting of results.

Finally, we want to improve dataset and model coverage. Although CIFAR-10 provides a useful testbed, community input should guide the addition of new datasets and model sizes, especially for settings considering generative AI applications. We welcome feedback and contributions from the community to help determine which of these directions would be most impactful in practice.

\newpage
\bibliographystyle{icml2025}
\bibliography{citations}

\appendix
\onecolumn
\section{Appendix}
\subsection{\KLOM{} sensitivity to number of compared models}
\label{app:klom_sensitivity}
\begin{figure}[H]
    \centering

    \begin{minipage}[b]{0.95\textwidth}
        \centering
        \includegraphics[height=4.5cm]{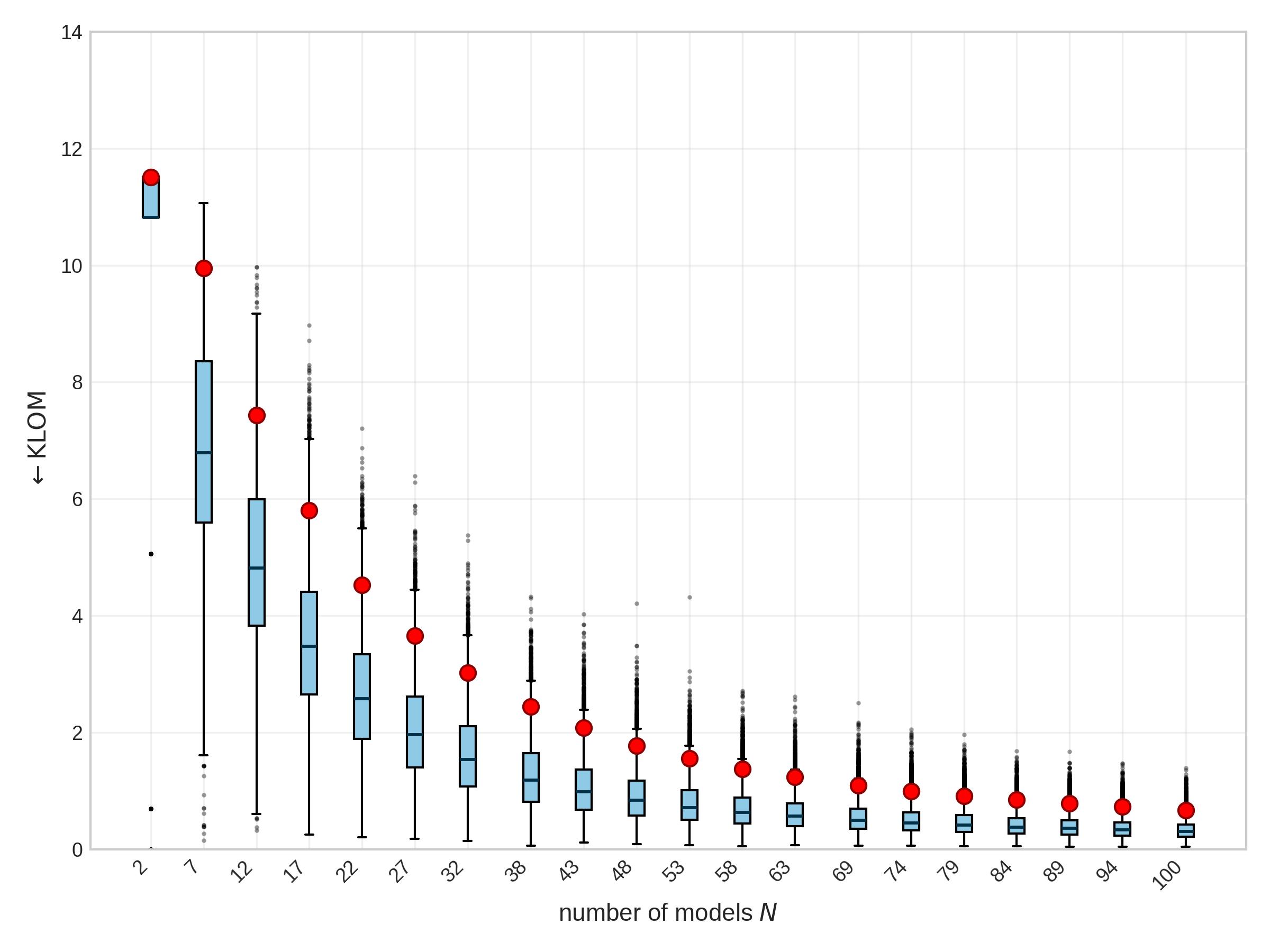}
        \captionof{subfigure}{Validation set.}
        \label{fig:klom_val}
    \end{minipage}

    \vspace{0.5em}

    \begin{minipage}[b]{0.95\textwidth}
        \centering
        \includegraphics[height=5.2cm]{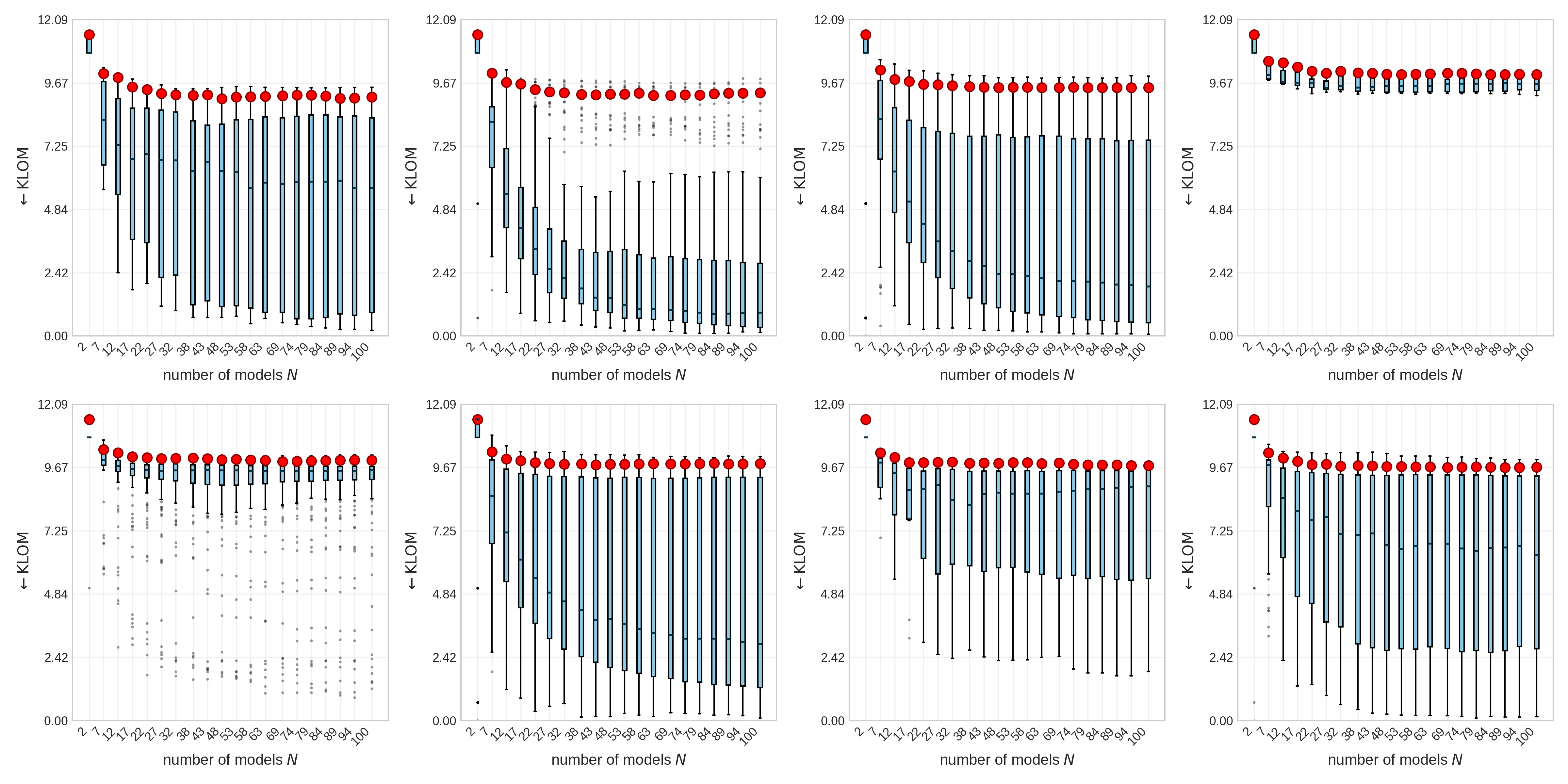}
        \captionof{subfigure}{Forget sets.}
        \label{fig:klom_fgt}
    \end{minipage}

    \vspace{0.5em}

    \begin{minipage}[b]{0.95\textwidth}
        \centering
        \includegraphics[height=5.2cm]{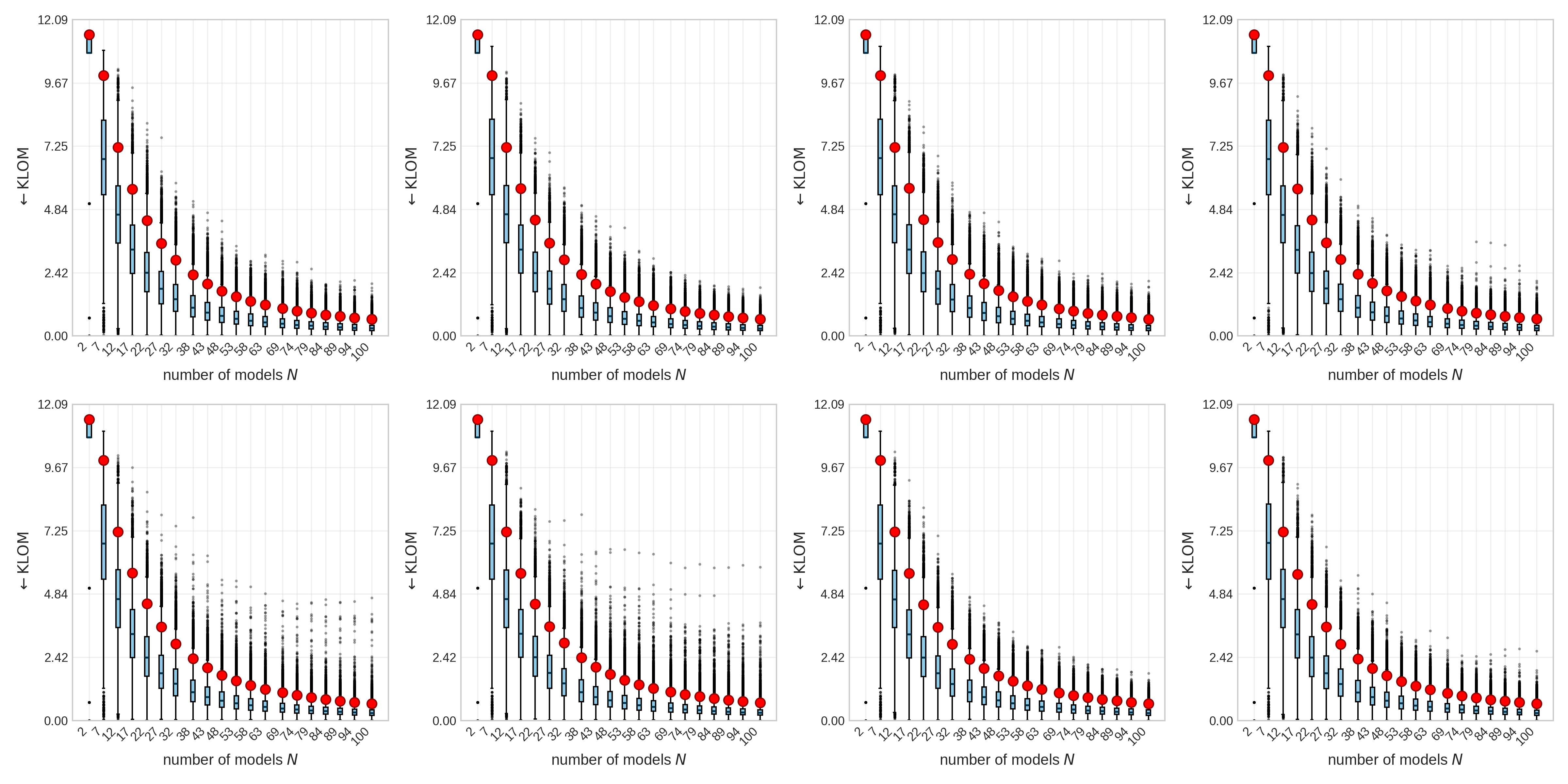}
        \captionof{subfigure}{Retain sets.}
        \label{fig:klom_ret}
    \end{minipage}

    \caption{
        \KLOM{} scores between pre-trained and oracle models scores on as a function of the number of compared models $N$ on CIFAR10 sets (1-8) \citep{georgiev2024attributetodeletemachineunlearningdatamodel}.
        The figure presents results for three data categories:
        (a) \textbf{Validation set}: a held-out test dataset, consistent across all forget configurations.
        (b) \textbf{Forget sets}: distributions for data points intended for unlearning.
        (c) \textbf{Retain sets}: distributions for data points to be preserved post-unlearning.
        In all panels, boxplots illustrate the \KLOM{} value distributions for $N$ ranging from 2 to 100. The red marker (\textcolor{red}{$\bullet$}) represents the 95-th percentile of \KLOM{} scores. Lower \KLOM{} values indicate better alignment of the pre-trained models with the oracle models and are expected in the Retain and Validation sets. We find $N=100$ to be sufficient for a reliable comparison.
    }
    \label{fig:klom_distribution}
\end{figure}

\subsection{Python API example}
\label{app:python_api}
The following snippet computes \KLOM{} from precomputed margins, without running any unlearning or requiring a GPU:

\begin{verbatim}
from easydub.data import ensure_dataset, load_margins_array
from easydub.eval import klom_from_margins
import numpy as np, torch

root = ensure_dataset()
n = 100
load = lambda kind, fid: torch.from_numpy(
    np.stack([load_margins_array(root, kind=kind, phase="val",
                                 forget_id=fid, model_id=i) for i in range(n)]))
pretrain = load("pretrain", None)
oracle   = load("oracle", 1)
score = klom_from_margins(oracle, pretrain)
\end{verbatim}

This computes the \KLOM{} baseline: how far the pretrain models are from the oracle models. A good unlearning method should produce a lower \KLOM{} than this baseline.

\end{document}